\pdfoutput=1
\documentclass[10pt,twocolumn,letterpaper]{article}

\usepackage[pagenumbers]{cvpr}

\usepackage{pifont}
\newcommand{\cmark}{\ding{51}}
\usepackage{colortbl}
\usepackage{arydshln}
\definecolor{bestcolor}{RGB}{0,90,40}
\newcommand{\best}[1]{\textcolor{bestcolor}{\textbf{#1}}}
\definecolor{gaingreen}{RGB}{0,90,40}
\newcommand{\gain}[1]{\,{\scriptsize\mdseries\color{gaingreen}[\textit{+#1}]}}
\newcommand{\gainbf}[1]{\,{\scriptsize\bfseries\color{gaingreen}[\textit{+#1}]}}
\newcommand{\loss}[1]{\,{\scriptsize\mdseries\color{black!45}[\textit{$-$#1}]}}
\newcommand{\gainrow}[1]{{\color{gaingreen}[\textit{+#1}]}}
\definecolor{sizeL}{RGB}{242,242,242}
\definecolor{sizeS}{RGB}{237,246,235}
\definecolor{sizeLmean}{RGB}{226,226,226}
\definecolor{sizeSmean}{RGB}{212,233,208}
\newcommand{\bestmark}[1]{\textcolor{bestcolor}{\textbf{#1}}}

\usepackage{multirow}

\AtBeginDocument{%
}

\definecolor{refcolor}{rgb}{0.80,0.33,0.00}
\definecolor{citecolor}{rgb}{0.10,0.50,0.25}
\AtBeginDocument{\hypersetup{linkcolor=refcolor,citecolor=citecolor,urlcolor=citecolor}}

\definecolor{cvprblue}{rgb}{0.21,0.49,0.74}
\usepackage[pagebackref,breaklinks,colorlinks,allcolors=cvprblue]{hyperref}

\def\confName{CVPR}
\def\confYear{2027}

\title{More Than Picturing the Future: Motion-Oriented Latent Reasoning\\for Generalizable Robot Manipulation}

\newcommand{\corrmark}{\mbox{\fontsize{10}{10}\selectfont\ding{41}}}
\author{
\textbf{Xin Zhang\textsuperscript{1}} \quad
\textbf{Yushuo Hou\textsuperscript{2,\,\corrmark}} \quad
\textbf{Anqi Zhao\textsuperscript{3}} \quad
\textbf{Moyun Liu\textsuperscript{2,\,\dag}} \\
\textbf{Xingwei Zhao\textsuperscript{2}} \quad
\textbf{Bo Tao\textsuperscript{2}} \quad
\textbf{Jiawei Du\textsuperscript{1}} \quad
\textbf{Joey Tianyi Zhou\textsuperscript{4,\,\corrmark}} \\[2pt]
\makebox[0pt][c]{\normalsize \textsuperscript{1}Institute of Advanced Intelligence and Computing, Agency for Science, Technology and Research (A*STAR), Singapore} \\
{\normalsize \textsuperscript{2}Huazhong University of Science and Technology, China} \\
{\normalsize \textsuperscript{3}University of Electronic Science and Technology of China, China} \\
{\normalsize \textsuperscript{4}Singapore Management University, Singapore} \\
{\normalsize \textsuperscript{\corrmark}Corresponding authors \quad \textsuperscript{\dag}Project lead} \\[2pt]
{\tt\small \{zhangx7, dujw\}@a-star.edu.sg, \{houyushuo, lmomoy, zhaoxingwei, taobo\}@hust.edu.cn} \\
{\tt\small zhaoanqiii@gmail.com, tyzhou@smu.edu.sg}
}

\begin{document}
\maketitle
\begingroup\setlength{\parskip}{0pt}
\begin{abstract}
Latent reasoning lets Vision-Language-Action (VLA) models deliberate at a fraction of the cost of explicit chain-of-thought, increasingly through visual foresight, which we show to be indispensable for predicting actions. Yet this paradigm is inherited from Vision-Language Models, where it serves understanding rather than action: each latent is aligned with an isolated future frame, often enriched with further modalities of that frame. We find that this binds the policy to the appearance of its training scenes, so that it degrades sharply once the scene changes. We argue that visual latents for action should capture not what a future frame looks like, but how the scene evolves toward the goal. We therefore propose \textbf{Motion-Oriented Latent Reasoning}, which redirects visual latents from the whole frame to the interaction that the action must bring about. Semantic latents, following the stages of embodied chain-of-thought, first establish what to accomplish and how. Conditioned on them, motion-oriented latents anticipate each future step from coarse to fine: from the global scene, to the interaction region spanned by the manipulated object and its destination, to the robot state. Relative to the state-of-the-art latent VLA LaST$_0$, our method raises the mean success rate on SimplerEnv-WidowX from 37.5\% to 55.2\% in distribution and from 22.4\% to 30.2\% under out-of-distribution shifts, and on an AgileX Piper dual-arm platform from 60.8\% to 90.8\% in training scenes and from 6.7\% to 37.5\% under environmental changes.
\end{abstract}
\par\endgroup
    
\begin{figure}[t]
  \centering
  \includegraphics[width=\linewidth]{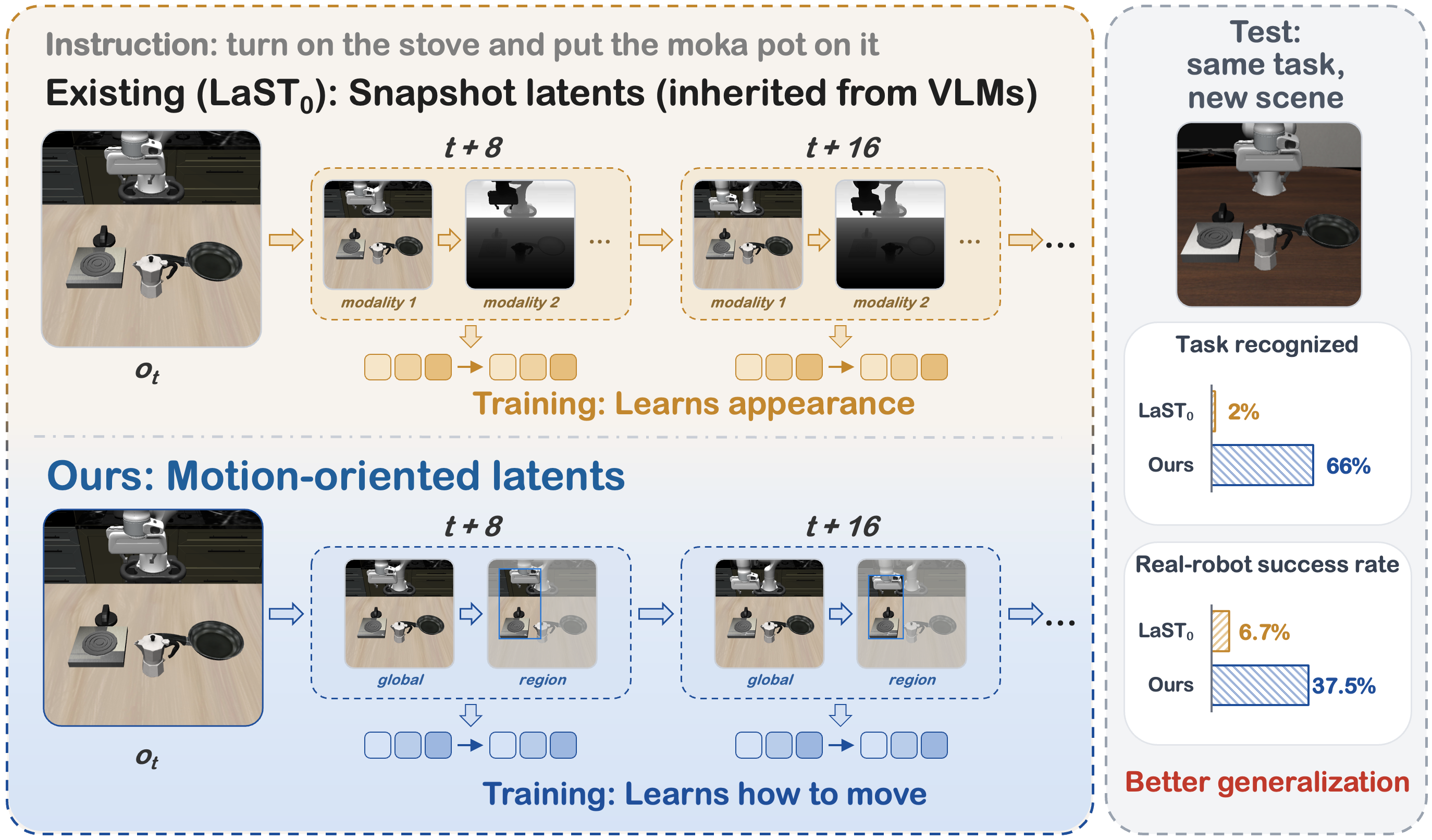}
  \vspace{-1.5em}
  \caption{Snapshot \vs motion-oriented latents. Existing latents learn what a future frame looks like; ours focus on the interaction region and learn how to move. Under a new background (right), LaST$_0$~\cite{liu2026last0} recognizes the task in 2\% of cases on LIBERO-Long (ours: 66\%), and under a likewise changed environment on a real robot, it succeeds in 6.7\% of trials (ours: 37.5\%).}
  \label{fig:teaser}
\vspace{-1.5em}
\end{figure}

\section{Introduction}
\label{sec:intro}

Building on pretrained Vision-Language Models (VLMs)~\cite{alayrac2022flamingo, karamcheti2024prismatic}, Vision-Language-Action (VLA) models~\cite{zitkovich2023rt2, kim2024openvla, black2024pi0} inherit web-scale visual-semantic priors and translate visual observations and language instructions directly into robot actions. Yet such reactive policies lack explicit deliberation over subgoals and action consequences, which is indispensable for complex, long-horizon manipulation~\cite{sun2025emmax, huang2026fastthinkact}. To bridge this gap, recent VLAs incorporate Chain-of-Thought (CoT)~\cite{wei2022chain} reasoning, verbalizing sub-task plans~\cite{zawalski2024ecot, sun2025emmax, huang2025thinkact, lin2025onetwovla} or anticipating future observations~\cite{zhao2025cotvla, zhang2025upvla, zhang2025dreamvla} before acting, and thereby attain stronger generalization.

\begin{figure}[t]
  \centering
  \includegraphics[width=0.9\linewidth]{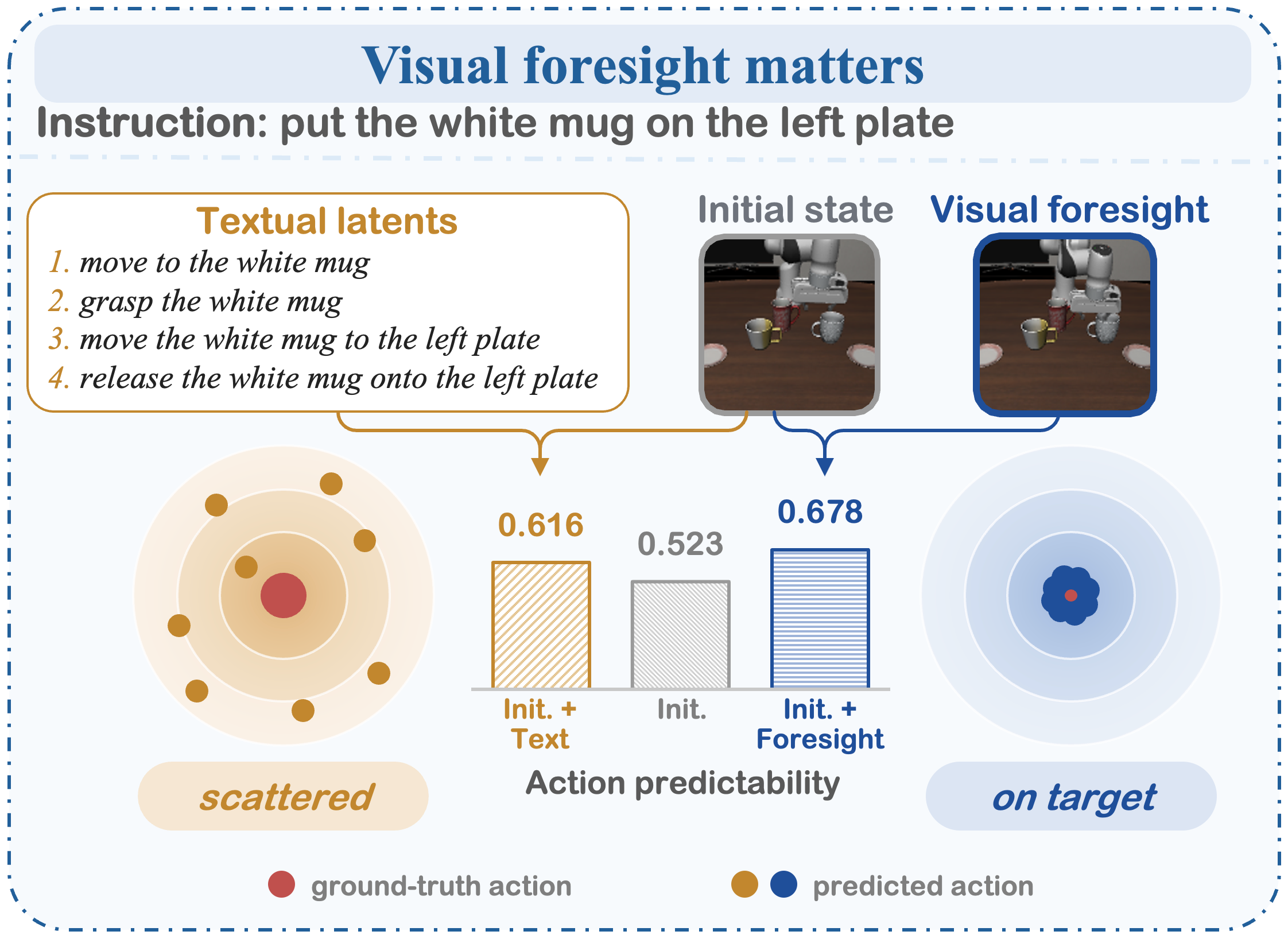}
  \caption{Visual foresight matters. Added to the initial state, textual latents leave the predicted actions scattered around the ground truth, whereas ground-truth visual foresight places them on target; bars report the action predictability of the initial state alone and with textual latents or visual foresight added.}
  \label{fig:foresight}
  \vspace{-1.5em}
\end{figure}

Despite these gains, explicit reasoning suffers from two inherent drawbacks: \textit{1)} autoregressive decoding incurs prohibitive latency~\cite{huang2026fastthinkact, bai2026laravla}, and \textit{2)} discretizing reasoning into tokens imposes an information bottleneck~\cite{liu2026last0, bai2026laravla}, both acute in embodied settings that demand real-time, continuous control. These limitations have spurred a shift toward \textit{Latent Reasoning}~\cite{map2025latent}, which unfolds reasoning in continuous hidden states~\cite{xu2026thinkinguncertainty} and endows compact latent tokens with task-relevant content, from verbalizable textual rationales in Heima~\cite{shen2025heima} to visual evidence such as trajectory maps in Mirage~\cite{yang2025machine}, regions of interest in LVR~\cite{li2025lvr}, 3D geometric priors in 3DThinker~\cite{chen2025think}, and perceptual maps in CoVT~\cite{qin2025chain}. A burgeoning body of work has brought this paradigm into VLAs~\cite{huang2025thinkact, ma2025hybrid, tan2026lcdrive, liu2026last0, xie2026flare, huang2026fastthinkact, bai2026laravla, luo2026lastvla}: Fast-ThinkAct~\cite{huang2026fastthinkact} distills lengthy textual reasoning into compact verbalizable latents, whereas LaST$_0$~\cite{liu2026last0}, LaRA-VLA~\cite{bai2026laravla}, and, in autonomous driving, LaST-VLA~\cite{luo2026lastvla} further ground their latents in future visual dynamics, 3D structure, and robot proprioception. Such grounding proves essential for embodied control: our oracle analysis reveals that, added to the current observation, ground-truth future visual evidence attains an action predictability\footnote{Action predictability measures how accurately the actions can be predicted from the given features; details are in \autoref{app:predictability}.} of 0.678, against 0.616 for textual latents (\autoref{fig:foresight}). \textit{Visual latents thus constitute an indispensable component of latent reasoning in VLAs.}

In VLMs, latent visual reasoning serves to comprehend the image at hand: its latents encode the current input across diverse modalities, from RGB features~\cite{yang2025machine, li2025lvr} and 3D geometry~\cite{chen2025think} to segmentation and depth maps~\cite{qin2025chain}, and the reasoning culminates in a single answer rather than an action sequence. Latent VLAs, however, inherit this understanding-oriented paradigm largely intact: each latent is aligned with the features of an isolated future frame, and richer foresight is pursued by stacking further modalities of that same frame~\cite{liu2026last0}. This raises a pivotal question: \emph{does a paradigm built to understand a frame suit a policy that must decide how to act?} We find that it does not. Reproducing what a future frame looks like rewards the backbone for encoding the appearance of the scene, and each additional modality of the same frame supplies further supervision on appearance rather than on how to move. For example, when only the workspace background changes while the task remains the same, the backbone of LaST$_0$~\cite{liu2026last0} recognizes the task in only 2\% of the observations (\autoref{fig:teaser}), mistaking a change of scene for a change of task. Bound to the appearance of its training scenes, the latent VLA overfits the training distribution: under the same kind of change on a real robot, it succeeds in only 6.7\% of trials, against 37.5\% for our method (\autoref{fig:teaser}).

In this paper, we argue that \textit{visual latents for action should capture not what a future frame looks like, but how the scene evolves toward the goal}. Textual reasoning unfolds step by step, with each step conditioned on its predecessors and the goal. Visual latents should likewise follow a logic of their own: within each future step, they should progress from the global scene to the interaction region and the robot state, \ie, from coarse to fine; across steps, each should build on the one before, so that successive steps chain into a motion path. Such a derivation supervises what changes, where, and in which order, which transfers across scenes, rather than the appearance that distinguishes one training scene from another. Based on this insight, we propose \textbf{Motion-Oriented Latent Reasoning}, which organizes latent reasoning as a logically structured derivation. Semantic latents, aligned with the semantic stages of embodied chain-of-thought~\cite{zawalski2024ecot} and trained under random textual prefix substitution, first establish what to accomplish and how. Conditioned on them, motion-oriented latents anticipate each future step from coarse to fine, zooming from the global scene into the interaction region spanned by the manipulated object and its destination before specifying the robot state. Without altering the inference procedure of latent reasoning, the policy thereby learns how to move rather than what the training scenes look like, and still recognizes the task in unseen scenes (\autoref{sec:exp_analysis}).

Our contributions are threefold: \textbf{(1)} We reveal that latent VLAs inherit an understanding-oriented paradigm of latent visual reasoning from VLMs, and show that aligning latents with isolated future frames and their modalities binds the policy to the appearance of its training scenes and impairs its generalization. \textbf{(2)} We propose Motion-Oriented Latent Reasoning, which organizes visual latents as a logically structured derivation, from coarse to fine within each step and chained across steps into a motion path, scaffolded by semantic latents. \textbf{(3)} Extensive experiments on SimplerEnv-WidowX, LIBERO, LIBERO-PRO, and an AgileX Piper dual-arm platform demonstrate that our method matches or improves in-distribution performance while substantially improving out-of-distribution generalization.

\begin{figure*}[t]
  \centering
  \includegraphics[width=\linewidth]{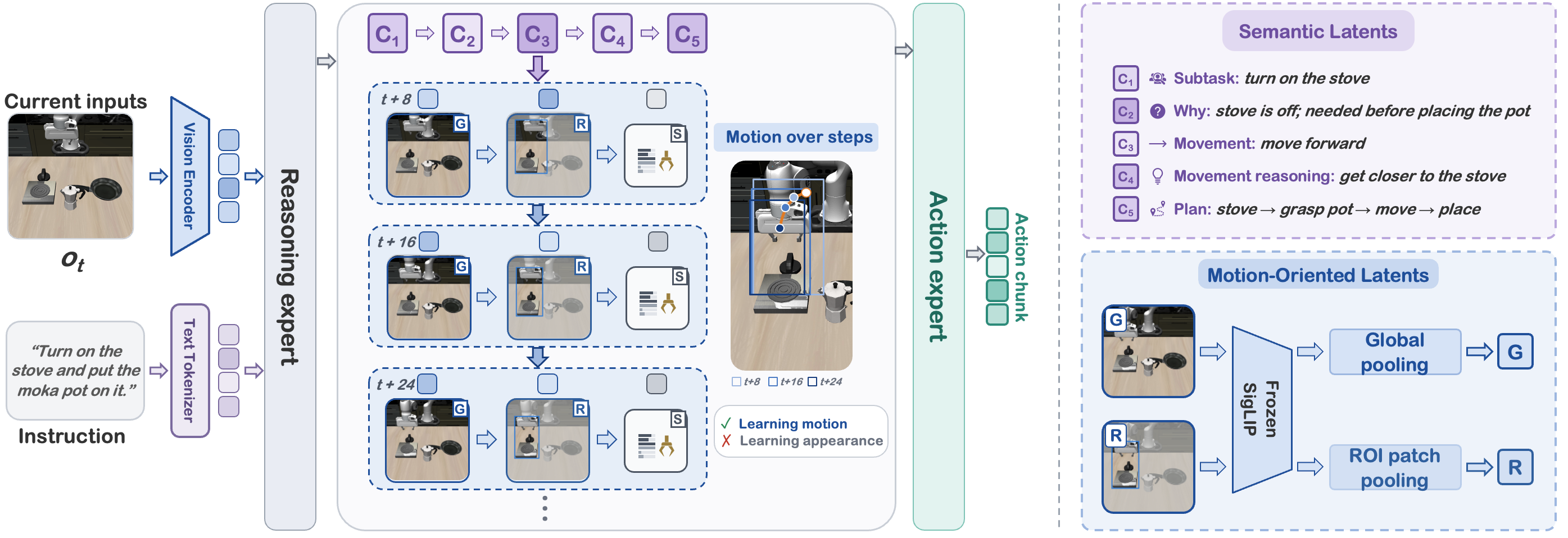}
  \caption{\textbf{Overview of Motion-Oriented Latent Reasoning.} \textbf{Left:} Given the current observation $o_t$ and the instruction, the reasoning expert first generates semantic latents $\mathbf{c}_1,\ldots,\mathbf{c}_5$ (C$_1$--C$_5$), which establish what to accomplish and how, and then motion-oriented latents that anticipate each future step from coarse to fine: the global scene (G), the interaction region (R), and the robot state (S). Each latent is conditioned on all its predecessors, and the action expert generates an action chunk conditioned on the observation, the instruction, and the latents. The inset (\textit{Motion over steps}) overlays the interaction regions at $t{+}8$, $t{+}16$, and $t{+}24$ with the end-effector path: the regions contract toward the destination and trace how the robot moves rather than what the scene looks like. \textbf{Top right:} Each semantic latent encodes one stage of embodied chain-of-thought. \textbf{Bottom right:} The global and regional targets pool future-frame features from the frozen SigLIP encoder over the whole frame and within the interaction region, respectively. All targets are used only during training.}
  \label{fig:framework}
\end{figure*}

\section{Related Work}
\label{sec:related}

\noindent\textbf{Reasoning in VLA models.}
Vision-Language-Action models adapt pretrained VLMs to robot control by mapping observations and instructions directly to actions~\cite{zitkovich2023rt2, kim2024openvla, octo2024, black2024pi0, li2024cogact, qu2025spatialvla, intelligence2025pi05}. To equip such reactive policies with deliberation, a line of work interposes explicit reasoning before acting: textual approaches verbalize subtasks, motions, and grounded object locations as embodied chain-of-thought~\cite{zawalski2024ecot, chen2025ecotlite, sun2025emmax, lin2025onetwovla}, while visual approaches synthesize a subgoal image~\cite{zhao2025cotvla, zhang2025upvla} or forecast dynamic regions, depth, and semantics~\cite{zhang2025dreamvla}. Others condition the policy on motion through predicted point tracks~\cite{wen2024atm, bharadhwaj2024track2act}, object flow~\cite{xu2024im2flow2act}, or tracked visual traces~\cite{zheng2025tracevla}, improving generalization across objects, scenes, and embodiments and corroborating that motion transfers where appearance does not. In both lines, however, the intermediate representation is decoded or tracked at every control step, incurring substantial latency at inference.

\noindent\textbf{Latent reasoning.}
Latent reasoning relocates deliberation into continuous hidden states~\cite{map2025latent}. In language models, the chain of thought is replaced by continuous thoughts~\cite{haotraining}, compressed into fewer latent steps~\cite{tan2025colar}, interleaved with discrete text tokens~\cite{su2025tokenassorted}, or condensed into verbalizable hidden tokens~\cite{shen2025heima}; VLMs extend it to visual evidence, aligning latent tokens with image features~\cite{yang2025machine}, regions of interest~\cite{li2025lvr}, 3D geometry~\cite{chen2025think}, or segmentation and depth cues~\cite{qin2025chain}, although recent analyses question how effectively such latents are exploited~\cite{li2026imagination, zhang2026unsilencing}. These latents serve to comprehend the image at hand rather than to act. In VLAs, RD-VLA~\cite{tur2026rdvla} scales test-time computation through latent iteration, Fast-ThinkAct~\cite{huang2026fastthinkact} compresses reasoning into verbalizable latents, and ThinkAct~\cite{huang2025thinkact} distills reinforced reasoning into a visual plan latent. Closer to our setting, implicit visual chain-of-thought~\cite{ma2025hybrid}, LaST$_0$~\cite{liu2026last0}, and LaRA-VLA~\cite{bai2026laravla} ground latents in future visual features, 3D structure, and robot proprioception, as do driving models that forecast future scenes in latent space~\cite{tan2026lcdrive, xie2026flare, luo2026lastvla, jia2026driveworldvla}. These methods inherit the understanding-oriented design of VLMs, supervising latents with snapshots of future frames or further modalities thereof, without organizing them around the interaction that the action must bring about.

\section{Method}
\label{sec:method}

\noindent\textbf{Problem Formulation.}
Consider a VLA policy built on a multimodal backbone $\Phi_{\theta}$, which, following LaST$_0$~\cite{liu2026last0}, is a Mixture-of-Transformers comprising a reasoning expert and an action expert that share a global self-attention context. Given an observation $o_t$ and a language instruction $\ell$ at time step $t$, $\Phi_{\theta}$ is required to predict an action chunk $\mathbf{A}_t=(\mathbf{a}_t,\ldots,\mathbf{a}_{t+H-1})$. The key insight of latent VLAs is to deliberate before acting by unfolding the reasoning in a sequence of continuous latents $\mathcal{Z}=(\mathbf{z}_1,\ldots,\mathbf{z}_K)$\footnote{Special tokens delimit the latent segment. Each latent is the last hidden state at the preceding position and is fed back as the input embedding of the next, so that $\mathbf{z}_k$ depends on $(o_t,\ell,\mathbf{z}_{<k})$.}, rather than in discrete textual tokens or generated images. Among them, visual latent methods, exemplified by LaST$_0$~\cite{liu2026last0}, optimize $\Phi_{\theta}$ with a joint objective that (i) \textbf{\textit{aligns each latent with a snapshot of the future}}, while (ii) \textbf{\textit{generating actions conditioned on the latents}},
\begin{equation}
  \theta^{*}=\arg\min_{\theta}\;
  \underbrace{\frac{1}{K}\sum_{k=1}^{K} d\big(\mathbf{z}_k,\bar{\mathbf{v}}_k\big)}_{\textit{\ding{172} visual latent alignment}}
  \;+\;
  \underbrace{\vphantom{\sum_{k=1}^{K}}\mathcal{L}_{\text{act}}\big(\mathbf{A}_t \mid o_t,\ell,\mathcal{Z}\big)}_{\textit{\ding{173} action generation}}.
  \label{eq:baseline}
\end{equation}
Here, $d(\mathbf{x},\mathbf{y})=1-\cos(\mathbf{x},\mathbf{y})$; $\bar{\mathbf{v}}_k$ denotes the patch-averaged visual feature of the future frame $I_{t+\Delta_k}$, alongside which further modalities of the same frame, such as point clouds, may be stacked; and $\mathcal{L}_{\text{act}}$ is the flow-matching objective~\cite{lipman2023flow} of the action expert.

Each target in \ding{172} thus summarizes what an entire future frame looks like. As the background and fixtures occupy most of the patches, this summary is dominated by the appearance of the scene rather than by the interaction that the action must bring about. Since the latents are produced by the same backbone that encodes the observation, matching these targets shapes the backbone itself and, as shown in \autoref{sec:intro}, binds it to the appearance of the training scenes.

\noindent\textbf{Overview.}
To learn how to move rather than what the scene looks like, we rethink what the latents should anticipate along two axes (\autoref{fig:framework}): (i) \textbf{\textit{where to look}}, shifting the visual targets from the entire frame to the interaction region spanned by the manipulated object and its destination, where the action takes effect; and (ii) \textbf{\textit{how to unfold}}, ordering the latents so that each builds on the ones before: semantic latents first establish what to accomplish and how; conditioned on them, the motion-oriented latents anticipate each future step from the global scene to the interaction region and the robot state, and successive steps connect into a motion path. Accordingly, the latent segment is laid out as
\begin{equation}
  \mathcal{Z}=\big(\underbrace{\mathbf{c}_1,\ldots,\mathbf{c}_M}_{\text{semantic latents}},\;
  \underbrace{\mathbf{g}_1,\boldsymbol{\rho}_1,\mathbf{s}_1,\;\ldots,\;\mathbf{g}_J,\boldsymbol{\rho}_J,\mathbf{s}_J}_{\text{motion-oriented latents}}\big),
  \label{eq:layout}
\end{equation}
with $M{=}5$ semantic latents and $J{=}4$ future steps, \ie, $K{=}M{+}3J{=}17$. Within the $j$-th step, the global latent $\mathbf{g}_j$ situates the step in the scene, the region latent $\boldsymbol{\rho}_j$ zooms into the interaction region, and the state latent $\mathbf{s}_j$ specifies the robot configuration; since each latent is conditioned on all its predecessors, successive region latents trace how the interaction unfolds over time. Inference is unchanged: the latents are generated as in \autoref{eq:baseline} and the action expert is untouched. \autoref{sec:reasoning_latents} and \autoref{sec:visual_latents} detail the semantic and motion-oriented latents, respectively.

\subsection{Semantic Latents as a Logical Scaffold}
\label{sec:reasoning_latents}

We derive the scaffold from embodied chain-of-thought annotations~\cite{zawalski2024ecot}, which decompose the reasoning at each step into $M$ semantic stages: \textit{Subtask}, \textit{Why}, \textit{Movement}, \textit{Movement reasoning}, and \textit{Plan}. The stages progress from the sub-goal and its rationale to the movement that realizes it and the plan that follows, so that each stage presupposes the ones before; each is represented by a semantic latent~$\mathbf{c}_m$.

\noindent\textbf{Textual prefix.}
Aligning each latent with its own stage alone would let the latents memorize their targets independently, without deriving one stage from another. Inspired by Token Assorted~\cite{su2025tokenassorted}, we therefore condition each latent on the annotations of the stages before it. For each training sample, we draw $\kappa\sim\mathcal{U}\{0,\ldots,M\}$ and replace the inputs at the first $\kappa$ semantic latent positions with $\bar{\mathbf{c}}_1,\ldots,\bar{\mathbf{c}}_{\kappa}$, where $\bar{\mathbf{c}}_m$ is the averaged token embedding of the annotation of the $m$-th stage (\eg, \textit{turn on the stove} for \textit{Subtask} in \autoref{fig:framework}). Since each latent is read out at the position preceding it, every $\mathbf{c}_m$ in this prefix ($m\le\kappa$) is predicted from the averaged annotation embeddings $\bar{\mathbf{c}}_1,\ldots,\bar{\mathbf{c}}_{m-1}$ of the preceding stages and is trained to match $\bar{\mathbf{c}}_m$,
\begin{equation}
  \mathcal{L}_{\text{sem}}=\frac{1}{\kappa}\sum_{m=1}^{\kappa} d\big(\mathbf{c}_m,\bar{\mathbf{c}}_m\big),
  \label{eq:reason_align}
\end{equation}
with $\mathcal{L}_{\text{sem}}{=}0$ when $\kappa{=}0$.

\noindent\textbf{Latent suffix.}
The remaining $M{-}\kappa$ stages receive no annotation embeddings: their positions keep placeholder inputs, and their latents have no target to align with. Following Heima~\cite{shen2025heima}, we instead require each of them to be verbalizable, \ie, taken alone as a prefix, $\mathbf{c}_m$ must suffice for the backbone to decode the annotation of its stage, supervised by the next-token cross-entropy $\mathcal{L}_{\text{dec}}$.

Since $\kappa$ is resampled for every sample, each stage falls sometimes in the prefix, where it must be inferred from the stages before it, and sometimes in the suffix, where it must carry its own content. The semantic latents thereby learn both the order of the stages and what each stage conveys. At inference, no annotations are provided and all semantic latents are generated by the backbone itself.

\subsection{Coarse-to-Fine Motion-Oriented Latents}
\label{sec:visual_latents}

Conditioned on the scaffold, the motion-oriented latents anticipate $J$ future steps $t{+}\Delta_j$ ($\Delta_j\in\{8,16,24,32\}$). Let $\mathbf{F}_j\in\mathbb{R}^{N\times D}$ denote the patch features of the future frame $I_{t+\Delta_j}$, extracted by the frozen visual encoder of the backbone and projected into its input space, where the $N$ patches form a regular grid over the image and $\mathbf{F}_{j,n}$ denotes the feature of the $n$-th patch. Each step is described by three latents of increasing specificity.

\noindent\textbf{Global latent.}
The global latent $\mathbf{g}_j$ captures the overall layout of the scene, including the arm, the objects, and their spatial arrangement. Its target $\bar{\mathbf{g}}_j=\frac{1}{N}\sum_{n=1}^{N}\mathbf{F}_{j,n}$ averages all patches and coincides with the snapshot target of \autoref{eq:baseline}, situating the step within the scene.

\noindent\textbf{Region latent.}
The region latent $\boldsymbol{\rho}_j$ anticipates the interaction region $\mathcal{B}_j$, the bounding box enclosing the manipulated object and its destination in the future frame; for operations without grasping, such as turning on a stove or opening a drawer, the gripper takes the place of the object. The component boxes are obtained offline with Grounding DINO~\cite{liu2024groundingdino} and SAM~3~\cite{carion2025sam3}, with manual correction where needed (\autoref{app:bbox}). The target averages the patches $\mathcal{P}_j$ that overlap $\mathcal{B}_j$,
\begin{equation}
  \bar{\boldsymbol{\rho}}_j=\frac{1}{|\mathcal{P}_j|}\sum_{n\in\mathcal{P}_j}\mathbf{F}_{j,n},
  \label{eq:region_target}
\end{equation}
and falls back to $\bar{\mathbf{g}}_j$ when no valid box is available, \eg, when detection fails or the future frame lies beyond the end of the episode. Computed only from the patches where the interaction takes place, $\bar{\boldsymbol{\rho}}_j$ is far less dominated by the background and distractors than $\bar{\mathbf{g}}_j$. Moreover, as the object approaches its destination, $\mathcal{B}_j$ shrinks and shifts across consecutive steps, so that the sequence of regional targets traces how the interaction progresses rather than what the scene looks like.

\noindent\textbf{State latent.}
The state latent $\mathbf{s}_j$ specifies the robot configuration that realizes the step. As in LaST$_0$~\cite{liu2026last0}, the future robot state is normalized, discretized into tokens, and embedded by the backbone, and the mean of these embeddings forms the target $\bar{\mathbf{s}}_j$.

\noindent\textbf{Objective.}
The motion-oriented latents are trained with
\begin{equation}
  \mathcal{L}_{\text{mot}}=\frac{1}{3J}\sum_{j=1}^{J}\Big[d\big(\mathbf{g}_j,\bar{\mathbf{g}}_j\big)+d\big(\boldsymbol{\rho}_j,\bar{\boldsymbol{\rho}}_j\big)+d\big(\mathbf{s}_j,\bar{\mathbf{s}}_j\big)\Big].
  \label{eq:visual_loss}
\end{equation}
Whereas the snapshot target of \autoref{eq:baseline} supervises only at the level of the whole frame, the supervision here descends from the scene to the interaction and further to the robot.

\begin{table}[t]
  \centering
  \caption{Success rates (\%) on SimplerEnv-WidowX in distribution (ID) and under the three out-of-distribution (OOD) splits of Interleave-VLA~\cite{fan2026interleave}: visual changes (Visual), novel objects (Novel Obj.), and novel categories (Novel Cat.). Numbers in brackets give the gain relative to LaST$_0$~\cite{liu2026last0}, and the best mean is in \protect\bestmark{bold}.}
  \vspace{-0.0em}
  \label{tab:simpler}
  \setlength{\tabcolsep}{3pt}
  \resizebox{\linewidth}{!}{%
  \begin{tabular}{l@{\hspace{4pt}}c@{\hspace{2pt}}*{12}{@{}>{\centering\arraybackslash}p{1.3em}@{}}@{\hspace{8pt}}c}
    \multicolumn{1}{l}{} & & & & & & & & & & & & & \\[-\normalbaselineskip]
    \toprule
    ID & Size & \multicolumn{3}{c}{Spoon} & \multicolumn{3}{c}{Carrot} & \multicolumn{3}{c}{Block} & \multicolumn{3}{c}{Eggplant} & Mean \\
    \midrule
    \rowcolor{sizeL} OpenVLA~\cite{kim2024openvla} & 7B & \multicolumn{3}{c}{4.2} & \multicolumn{3}{c}{12.5} & \multicolumn{3}{c}{8.3} & \multicolumn{3}{c}{16.7} & 10.4 \\
    \rowcolor{sizeL} OpenVLA-OFT~\cite{kim2025oft} & 7B & \multicolumn{3}{c}{16.0} & \multicolumn{3}{c}{18.0} & \multicolumn{3}{c}{0.0} & \multicolumn{3}{c}{58.0} & 23.0 \\
    \rowcolor{sizeL} MAPS~\cite{huang2026maps} & 7B & \multicolumn{3}{c}{2.0} & \multicolumn{3}{c}{16.0} & \multicolumn{3}{c}{0.0} & \multicolumn{3}{c}{74.0} & 23.0 \\
    \noalign{\vskip 0.6pt}\hdashline\noalign{\vskip 0.6pt}
    \rowcolor{sizeL} SpatialVLA~\cite{qu2025spatialvla} & 4B & \multicolumn{3}{c}{29.2} & \multicolumn{3}{c}{33.3} & \multicolumn{3}{c}{33.3} & \multicolumn{3}{c}{100.0} & 49.0 \\
    \noalign{\vskip 0.6pt}\hdashline\noalign{\vskip 0.6pt}
    \rowcolor{sizeL} Octo-Base~\cite{octo2024} & 0.09B & \multicolumn{3}{c}{20.8} & \multicolumn{3}{c}{20.8} & \multicolumn{3}{c}{0.0} & \multicolumn{3}{c}{45.8} & 21.9 \\
    \midrule
    \rowcolor{sizeS} LaST$_0$~\cite{liu2026last0} & 3.3B & \multicolumn{3}{c}{25.0} & \multicolumn{3}{c}{25.0} & \multicolumn{3}{c}{16.7} & \multicolumn{3}{c}{83.3} & 37.5 \\
    \rowcolor{sizeS} \textbf{Ours} & 3.3B & \multicolumn{3}{c}{58.3} & \multicolumn{3}{c}{29.2} & \multicolumn{3}{c}{37.5} & \multicolumn{3}{c}{95.8} & \cellcolor{sizeSmean}\best{55.2} \\[-0.5pt]
    \rowcolor{sizeS} & & \multicolumn{3}{c}{\gainrow{33.3}} & \multicolumn{3}{c}{\gainrow{4.2}} & \multicolumn{3}{c}{\gainrow{20.8}} & \multicolumn{3}{c}{\gainrow{12.5}} & \cellcolor{sizeSmean}\gainrow{17.7} \\
    \midrule[\heavyrulewidth]
    OOD & Size & \multicolumn{4}{c}{Visual} & \multicolumn{4}{c}{Novel Obj.} & \multicolumn{4}{c}{Novel Cat.} & Mean \\
    \midrule
    \rowcolor{sizeL} OpenVLA~\cite{kim2024openvla} & 7B & \multicolumn{4}{c}{6.0} & \multicolumn{4}{c}{15.3} & \multicolumn{4}{c}{0.7} & 7.3 \\
    \rowcolor{sizeL} OpenVLA-OFT~\cite{kim2025oft} & 7B & \multicolumn{4}{c}{16.0} & \multicolumn{4}{c}{9.3} & \multicolumn{4}{c}{0.7} & 8.7 \\
    \rowcolor{sizeL} MAPS~\cite{huang2026maps} & 7B & \multicolumn{4}{c}{18.7} & \multicolumn{4}{c}{20.0} & \multicolumn{4}{c}{12.7} & 17.1 \\
    \noalign{\vskip 0.6pt}\hdashline\noalign{\vskip 0.6pt}
    \rowcolor{sizeL} SpatialVLA~\cite{qu2025spatialvla} & 4B & \multicolumn{4}{c}{29.3} & \multicolumn{4}{c}{20.0} & \multicolumn{4}{c}{22.0} & 23.8 \\
    \noalign{\vskip 0.6pt}\hdashline\noalign{\vskip 0.6pt}
    \rowcolor{sizeL} Octo-Base~\cite{octo2024} & 0.09B & \multicolumn{4}{c}{7.3} & \multicolumn{4}{c}{0.7} & \multicolumn{4}{c}{1.3} & 3.1 \\
    \midrule
    \rowcolor{sizeS} LaST$_0$~\cite{liu2026last0} & 3.3B & \multicolumn{4}{c}{18.7} & \multicolumn{4}{c}{22.0} & \multicolumn{4}{c}{26.7} & 22.4 \\
    \rowcolor{sizeS} \textbf{Ours} & 3.3B & \multicolumn{4}{c}{32.0} & \multicolumn{4}{c}{30.0} & \multicolumn{4}{c}{28.7} & \cellcolor{sizeSmean}\best{30.2} \\[-0.5pt]
    \rowcolor{sizeS} & & \multicolumn{4}{c}{\gainrow{13.3}} & \multicolumn{4}{c}{\gainrow{8.0}} & \multicolumn{4}{c}{\gainrow{2.0}} & \cellcolor{sizeSmean}\gainrow{7.8} \\
    \bottomrule
  \end{tabular}}
 \vspace{-1em}
\end{table}

\subsection{Training and Inference}
\label{sec:training}

\noindent\textbf{Two-pass training.}
Following LaST$_0$~\cite{liu2026last0}, training proceeds in two forward passes over the same sequence. In the first pass, the inputs at the motion-oriented latent positions are replaced by their targets, and those at the first $\kappa$ semantic latent positions by $\bar{\mathbf{c}}_1,\ldots,\bar{\mathbf{c}}_{\kappa}$. The remaining $M{-}\kappa$ semantic latent positions keep their placeholder inputs. Each latent is read out from the hidden state at the preceding position, and $\mathcal{L}_{\text{sem}}$ (\autoref{eq:reason_align}), $\mathcal{L}_{\text{dec}}$, and $\mathcal{L}_{\text{mot}}$ (\autoref{eq:visual_loss}) are computed on these predictions. In the second pass, all $K$ latent positions are filled with the latents predicted in the first pass, and the action expert is optimized with $\mathcal{L}_{\text{act}}$ conditioned on them. The action expert is thus trained on the latents that the backbone itself produces, as at inference, and the action loss is back-propagated through these latents into the first pass. The overall objective is
\begin{equation}
  \mathcal{L}=\mathcal{L}_{\text{act}}+\mathcal{L}_{\text{mot}}+\lambda_{\text{sem}}\,\mathcal{L}_{\text{sem}}+\lambda_{\text{dec}}\,\mathcal{L}_{\text{dec}}.
  \label{eq:total}
\end{equation}

\noindent\textbf{Inference.}
The text annotations, bounding boxes, and future frames serve only to construct the training targets. At inference, all $K$ latents are generated autoregressively in the continuous hidden space without decoding any text, after which the action expert produces the action chunk. The inference procedure thus follows that of LaST$_0$~\cite{liu2026last0}, preserving the efficiency of latent reasoning.

\section{Experiments}
\label{sec:exp}

In this section, we first evaluate our method in simulation in terms of success rate and generalization (\autoref{sec:exp_sim}). We then ablate each component and examine whether our latents resolve the failure identified in \autoref{sec:intro} (\autoref{sec:exp_ablation}), and finally validate our method in the real world (\autoref{sec:exp_real}).

\begin{table*}[t]
  \stepcounter{table}
  \centering
  \caption{Latent ablation: success rates (\%) in distribution (ID) on LIBERO and out of distribution (OOD) on the Spatial suite of LIBERO-PRO~\cite{zhou2025liberopro}. \cmark~marks the latent groups that are retained, \# is the total number of latents, the numbers in brackets give the gain over the baseline without latents, and the best mean is in \protect\bestmark{bold}.}
  \vspace{-0.5em}
  \label{tab:latent_ablation}
  \setlength{\tabcolsep}{4pt}
  \resizebox{\ifdim\width>\textwidth\textwidth\else\width\fi}{!}{%
  \begin{tabular}{ccc*{5}{c}*{6}{c}}
    \toprule
    \multicolumn{3}{c}{Latents} & \multicolumn{5}{c}{ID} & \multicolumn{6}{c}{OOD (Pro)} \\
    \cmidrule(lr){1-3} \cmidrule(lr){4-8} \cmidrule(lr){9-14}
    Semantic & Motion & \# & Spatial & Object & Goal & Long & Mean & Obj. & Swap & Lang. & Task & Env. & Mean \\
    \midrule
    & & 0
      & 81.8 & 99.2 & 41.2 & 86.8 & 77.3 & 68.2 & 7.0 & 81.0 & 22.2 & 25.6 & 40.8 \\
    & \cmark & 12
      & 90.0\gain{8.2} & 99.2\gain{0.0} & 91.2\gain{50.0} & 85.2\loss{1.6} & 91.4\gain{14.1} & 78.4\gain{10.2} & 17.4\gain{10.4} & 91.4\gain{10.4} & 29.0\gain{6.8} & 35.0\gain{9.4} & 50.2\gain{9.4} \\
    \cmark & \cmark & 17
      & 99.2\gain{17.4} & 99.8\gain{0.6} & 96.0\gain{54.8} & 93.6\gain{6.8} & \cellcolor{sizeSmean}\best{97.2}\gainbf{19.9} & 89.8\gain{21.6} & 18.2\gain{11.2} & 89.2\gain{8.2} & 36.4\gain{14.2} & 32.8\gain{7.2} & \cellcolor{sizeSmean}\best{53.3}\gainbf{12.5} \\
    \bottomrule
  \end{tabular}}
   \vspace{-1em}
\end{table*}
\addtocounter{table}{-2}

\suppressfloats[t]%
\begin{table}[t]
  \centering
  \caption{Success rates (\%) on LIBERO in distribution (ID) and out of distribution on the Spatial suite. OOD reports the five perturbations of LIBERO-PRO~\cite{zhou2025liberopro}: manipulated objects (Obj.), swapped initial positions (Swap), paraphrased instructions (Lang.), modified tasks (Task), and environments (Env.). Numbers in brackets give the gain relative to LaST$_0$~\cite{liu2026last0}, and the best mean is in \protect\bestmark{bold}.}
  \vspace{-0.0em}
  \label{tab:libero}
  \setlength{\tabcolsep}{3pt}
  \resizebox{\linewidth}{!}{%
  \begin{tabular}{l@{\hspace{4pt}}c@{\hspace{2pt}}*{30}{@{}>{\centering\arraybackslash}p{0.8em}@{}}}
    \multicolumn{1}{l}{} & & & & & & & & & & & & & & & & & & & & & & & & & & & & & & & \\[-\normalbaselineskip]
    \toprule
    ID & Size & \multicolumn{6}{c}{Spatial} & \multicolumn{6}{c}{Object} & \multicolumn{6}{c}{Goal} & \multicolumn{6}{c}{Long} & \multicolumn{6}{c}{Mean} \\
    \midrule
    \rowcolor{sizeL} OpenVLA~\cite{kim2024openvla} & 7B & \multicolumn{6}{c}{84.7} & \multicolumn{6}{c}{88.4} & \multicolumn{6}{c}{79.2} & \multicolumn{6}{c}{53.7} & \multicolumn{6}{c}{76.5} \\
    \rowcolor{sizeL} CogACT~\cite{li2024cogact} & 7B & \multicolumn{6}{c}{97.2} & \multicolumn{6}{c}{98.0} & \multicolumn{6}{c}{90.2} & \multicolumn{6}{c}{88.8} & \multicolumn{6}{c}{93.6} \\
    \rowcolor{sizeL} CoT-VLA~\cite{zhao2025cotvla} & 7B & \multicolumn{6}{c}{87.5} & \multicolumn{6}{c}{91.6} & \multicolumn{6}{c}{87.6} & \multicolumn{6}{c}{69.0} & \multicolumn{6}{c}{83.9} \\
    \rowcolor{sizeL} OpenVLA-OFT~\cite{kim2025oft} & 7B & \multicolumn{6}{c}{97.6} & \multicolumn{6}{c}{98.4} & \multicolumn{6}{c}{97.9} & \multicolumn{6}{c}{94.5} & \multicolumn{6}{c}{97.1} \\
    \rowcolor{sizeL} MolmoAct~\cite{lee2025molmoact} & 7B & \multicolumn{6}{c}{87.0} & \multicolumn{6}{c}{95.4} & \multicolumn{6}{c}{87.6} & \multicolumn{6}{c}{77.2} & \multicolumn{6}{c}{86.8} \\
    \noalign{\vskip 0.6pt}\hdashline\noalign{\vskip 0.6pt}
    \rowcolor{sizeL} SpatialVLA~\cite{qu2025spatialvla} & 4B & \multicolumn{6}{c}{88.2} & \multicolumn{6}{c}{89.9} & \multicolumn{6}{c}{78.6} & \multicolumn{6}{c}{55.5} & \multicolumn{6}{c}{78.1} \\
    \noalign{\vskip 0.6pt}\hdashline\noalign{\vskip 0.6pt}
    \rowcolor{sizeL} $\pi_0$~\cite{black2024pi0} & 3.3B & \multicolumn{6}{c}{96.8} & \multicolumn{6}{c}{98.8} & \multicolumn{6}{c}{95.8} & \multicolumn{6}{c}{85.2} & \multicolumn{6}{c}{94.2} \\
    \rowcolor{sizeL} $\pi_{0.5}$~\cite{intelligence2025pi05} & 3.3B & \multicolumn{6}{c}{98.8} & \multicolumn{6}{c}{98.2} & \multicolumn{6}{c}{98.0} & \multicolumn{6}{c}{92.4} & \multicolumn{6}{c}{96.9} \\
    \rowcolor{sizeL} NORA-Long~\cite{hung2025nora} & 3B & \multicolumn{6}{c}{92.2} & \multicolumn{6}{c}{95.4} & \multicolumn{6}{c}{89.4} & \multicolumn{6}{c}{74.6} & \multicolumn{6}{c}{87.9} \\
    \midrule
    \rowcolor{sizeS} LaST$_0$~\cite{liu2026last0} & 3.3B & \multicolumn{6}{c}{98.4} & \multicolumn{6}{c}{99.4} & \multicolumn{6}{c}{95.8} & \multicolumn{6}{c}{92.0} & \multicolumn{6}{c}{96.4} \\
    \rowcolor{sizeS} \textbf{Ours} & 3.3B & \multicolumn{6}{c}{99.2} & \multicolumn{6}{c}{99.8} & \multicolumn{6}{c}{96.0} & \multicolumn{6}{c}{93.6} & \multicolumn{6}{c}{\cellcolor{sizeSmean}\best{97.2}} \\[-0.5pt]
    \rowcolor{sizeS} & & \multicolumn{6}{c}{\gainrow{0.8}} & \multicolumn{6}{c}{\gainrow{0.4}} & \multicolumn{6}{c}{\gainrow{0.2}} & \multicolumn{6}{c}{\gainrow{1.6}} & \multicolumn{6}{c}{\cellcolor{sizeSmean}\gainrow{0.8}} \\
    \midrule[\heavyrulewidth]
    OOD (Pro) & Size & \multicolumn{5}{c}{Obj.} & \multicolumn{5}{c}{Swap} & \multicolumn{5}{c}{Lang.} & \multicolumn{5}{c}{Task} & \multicolumn{5}{c}{Env.} & \multicolumn{5}{c}{Mean} \\
    \midrule
    \rowcolor{sizeL} OpenVLA-OFT~\cite{kim2025oft} & 7B & \multicolumn{5}{c}{97.0} & \multicolumn{5}{c}{0.0} & \multicolumn{5}{c}{97.0} & \multicolumn{5}{c}{0.0} & \multicolumn{5}{c}{89.0} & \multicolumn{5}{c}{\cellcolor{sizeLmean}\textbf{56.6}} \\
    \rowcolor{sizeL} MolmoAct~\cite{lee2025molmoact} & 7B & \multicolumn{5}{c}{90.0} & \multicolumn{5}{c}{0.0} & \multicolumn{5}{c}{88.0} & \multicolumn{5}{c}{0.0} & \multicolumn{5}{c}{--} & \multicolumn{5}{c}{44.5} \\
    \noalign{\vskip 0.6pt}\hdashline\noalign{\vskip 0.6pt}
    \rowcolor{sizeL} $\pi_0$~\cite{black2024pi0} & 3.3B & \multicolumn{5}{c}{95.0} & \multicolumn{5}{c}{0.0} & \multicolumn{5}{c}{97.0} & \multicolumn{5}{c}{0.0} & \multicolumn{5}{c}{60.0} & \multicolumn{5}{c}{50.4} \\
    \rowcolor{sizeL} $\pi_{0.5}$~\cite{intelligence2025pi05} & 3.3B & \multicolumn{5}{c}{97.0} & \multicolumn{5}{c}{20.0} & \multicolumn{5}{c}{97.0} & \multicolumn{5}{c}{1.0} & \multicolumn{5}{c}{46.0} & \multicolumn{5}{c}{52.2} \\
    \rowcolor{sizeL} NORA-Long~\cite{hung2025nora} & 3B & \multicolumn{5}{c}{92.0} & \multicolumn{5}{c}{0.0} & \multicolumn{5}{c}{91.0} & \multicolumn{5}{c}{0.0} & \multicolumn{5}{c}{--} & \multicolumn{5}{c}{45.8} \\
    \midrule
    \rowcolor{sizeS} LaST$_0$~\cite{liu2026last0} & 3.3B & \multicolumn{5}{c}{85.0} & \multicolumn{5}{c}{10.8} & \multicolumn{5}{c}{87.2} & \multicolumn{5}{c}{32.8} & \multicolumn{5}{c}{29.8} & \multicolumn{5}{c}{49.1} \\
    \rowcolor{sizeS} \textbf{Ours} & 3.3B & \multicolumn{5}{c}{89.8} & \multicolumn{5}{c}{18.2} & \multicolumn{5}{c}{89.2} & \multicolumn{5}{c}{36.4} & \multicolumn{5}{c}{32.8} & \multicolumn{5}{c}{53.3} \\[-0.5pt]
    \rowcolor{sizeS} & & \multicolumn{5}{c}{\gainrow{4.8}} & \multicolumn{5}{c}{\gainrow{7.4}} & \multicolumn{5}{c}{\gainrow{2.0}} & \multicolumn{5}{c}{\gainrow{3.6}} & \multicolumn{5}{c}{\gainrow{3.0}} & \multicolumn{5}{c}{\gainrow{4.2}} \\
    \bottomrule
  \end{tabular}}
 \vspace{-1em}
\end{table}
\stepcounter{table}%

\subsection{Simulation Experiments}
\label{sec:exp_sim}

\noindent\textbf{Benchmark setup.}
We evaluate on SimplerEnv-WidowX~\cite{li2024simpler} and LIBERO~\cite{liu2023libero}, each paired with an out-of-distribution (OOD) counterpart. SimplerEnv-WidowX evaluates policies trained on BridgeData~V2~\cite{walke2023bridgedata} in four real-to-sim tasks, \ie, placing a spoon on a towel, a carrot on a plate, stacking blocks, and putting an eggplant in a basket. For its OOD evaluation, we adopt the extended suite of Interleave-VLA~\cite{fan2026interleave} with the task list of MAPS~\cite{huang2026maps}, comprising nine tasks in three splits: Visual alters the lighting, table texture, and background; Novel Object introduces unseen instances of familiar categories; and Novel Category introduces entirely unseen categories. LIBERO comprises four suites, \ie, Spatial, Object, Goal, and Long, each with 500 demonstrations over 10 tasks. For its OOD evaluation, we adopt LIBERO-PRO~\cite{zhou2025liberopro}, which perturbs the same suites in five ways: the manipulated objects, their initial positions, the instructions, the tasks, and the environments.

\noindent\textbf{Training and evaluation protocol.}
Our model is built on Janus-Pro-1B~\cite{chen2025januspro} and initialized from the action expert pretrained by LaST$_0$~\cite{liu2026last0}, which generates 16-step action chunks via flow matching~\cite{lipman2023flow}. Embodied chain-of-thought annotations follow~\cite{zawalski2024ecot}, and the interaction regions are obtained offline with Grounding DINO~\cite{liu2024groundingdino} and SAM~3~\cite{carion2025sam3} (\autoref{app:bbox}). We train on 8 NVIDIA H200 GPUs with AdamW at a total batch size of 128 and a learning rate of $10^{-4}$ with cosine decay, and set $\lambda_{\text{sem}}{=}0.05$ and $\lambda_{\text{dec}}{=}0.01$ in \autoref{eq:total}. The baselines are listed in \autoref{tab:simpler} and \autoref{tab:libero}. On SimplerEnv-WidowX, each in-distribution task is evaluated over 24 trials and each OOD task over 50 trials. On LIBERO, our model is evaluated over 500 trials per suite, and on LIBERO-PRO, over 500 trials per perturbation of the Spatial suite. On both benchmarks, all methods are evaluated under the same setting.

\noindent\textbf{SimplerEnv-WidowX results.}
As shown in \autoref{tab:simpler}, our method attains the highest mean success rate both in and out of distribution. In distribution, it reaches 55.2\%, exceeding LaST$_0$~\cite{liu2026last0}, on which it builds, by 17.7 points and the strongest baseline, SpatialVLA, by 6.2 points; on Spoon, it attains 58.3\%, twice the best competing result. Out of distribution, it reaches 30.2\%, against 23.8\% for SpatialVLA and 22.4\% for LaST$_0$~\cite{liu2026last0}, and leads all methods on every split: Visual (32.0\%), Novel Object (30.0\%), and Novel Category (28.7\%), improving on LaST$_0$~\cite{liu2026last0} by 13.3, 8.0, and 2.0 points, respectively. It does so with 3.3B parameters, fewer than SpatialVLA (4B) and OpenVLA (7B).

\noindent\textbf{LIBERO benchmark results.}
As shown in \autoref{tab:libero}, our method attains the highest in-distribution mean of 97.2\%, ahead of OpenVLA-OFT (97.1\%) and $\pi_{0.5}$ (96.9\%), with the best results on Spatial (99.2\%) and Object (99.8\%). It improves on LaST$_0$~\cite{liu2026last0} in every suite, most notably by 1.6 points on the multi-stage Long suite. On the Spatial suite of LIBERO-PRO, it improves on LaST$_0$~\cite{liu2026last0} under all five perturbations, raising the mean from 49.1\% to 53.3\%, with the largest gains under swapped positions and manipulated objects (7.4 and 4.8 points). Our method thus surpasses $\pi_{0.5}$ and OpenVLA-OFT in distribution; out of distribution, it exceeds $\pi_{0.5}$ (52.2\%) while trailing OpenVLA-OFT (56.6\%).

\begin{figure}[t]
  \centering
  \includegraphics[width=\linewidth]{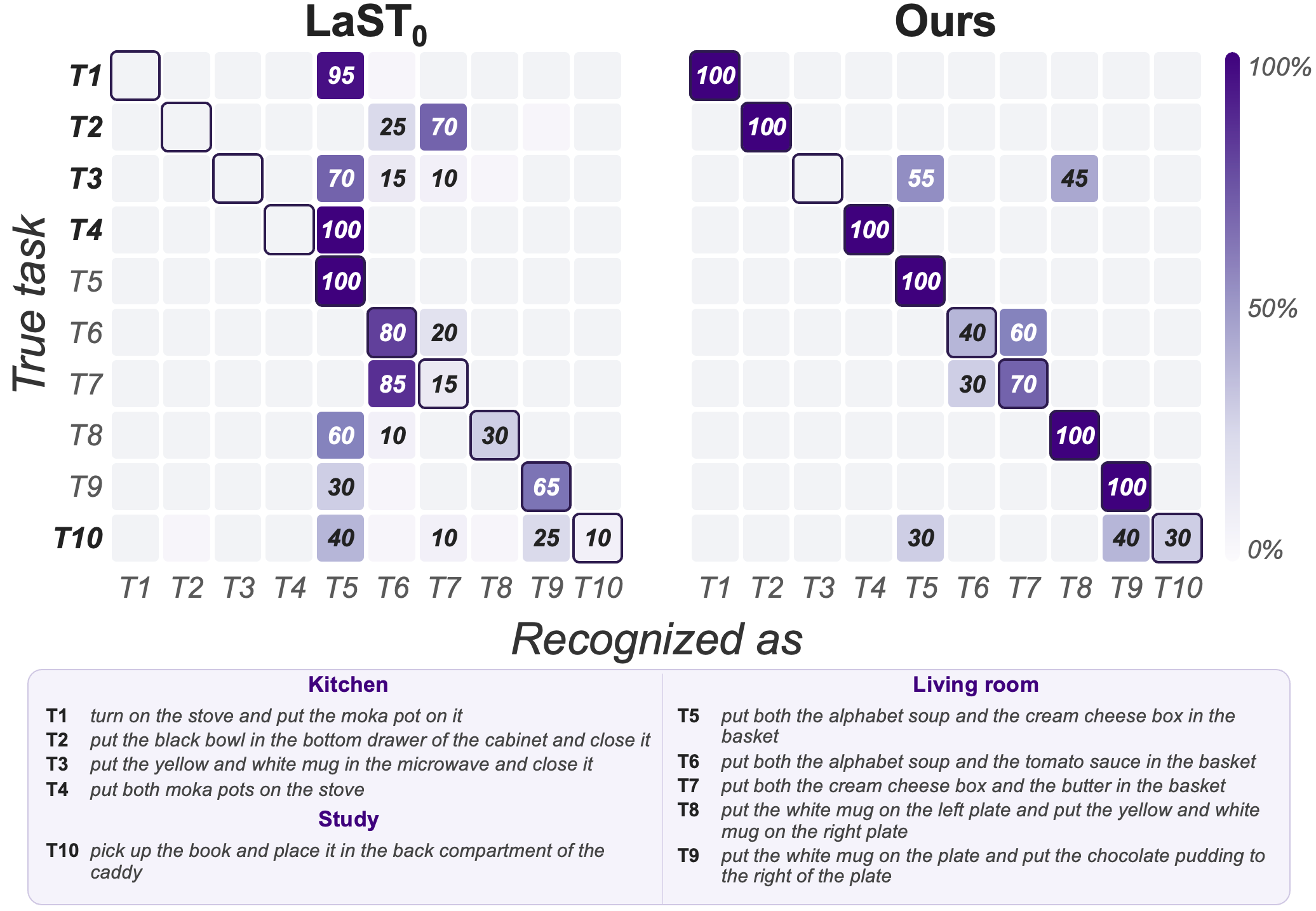}
  \vspace{-1.0em}
  \caption{Task recognition from the backbone under a new background on LIBERO-Long. Each row gives the fraction of perturbed observations of a task assigned to each original-scene task; bold tasks are those whose scene visibly changes.}
  \vspace{-1.5em}
  \label{fig:task_confusion}
\end{figure}

\subsection{Ablation Study and Analysis}
\label{sec:exp_ablation}

\noindent\textbf{What does each latent group contribute?}
\autoref{tab:latent_ablation} ablates the two latent groups against a baseline whose latents are removed at inference, in distribution on LIBERO and out of distribution on the Spatial suite of LIBERO-PRO. The twelve motion-oriented latents alone raise the in-distribution mean from 77.3\% to 91.4\%; the gain concentrates on Goal, which rises from 41.2\% to 91.2\%. Out of distribution, they improve all five perturbations, by 6.8 to 10.4 points, raising the mean from 40.8\% to 50.2\%. Adding the five semantic latents improves every suite in distribution, most notably Spatial and Long (by 9.2 and 8.4 points), lifting the mean to 97.2\%. Out of distribution, it further raises manipulated objects from 78.4\% to 89.8\% and modified tasks from 29.0\% to 36.4\%, lifting the mean to 53.3\%. The two groups are thus complementary: motion-oriented latents furnish most of the gain both in and out of distribution, whereas the semantic scaffold extends it to the long-horizon suite and to changes of the manipulated object and the task.

\phantomsection\label{sec:exp_analysis}
\noindent\textbf{Does the backbone recognize the task under a new scene?}
We probe whether the training target determines what the backbone encodes. On the ten LIBERO-Long tasks, each rendered from 20 initial states in the original scene and under the environmental perturbation of LIBERO-PRO~\cite{zhou2025liberopro}, we take the hidden states of the third-person image tokens at the middle layer of the backbone. As these tokens precede the latent segment under causal attention, their states are independent of the generated latents and reflect only what training has instilled. Each perturbed observation is assigned to the task whose original-scene centroid is nearest. On the five tasks whose scene visibly changes, the backbone of LaST$_0$~\cite{liu2026last0} recognizes the task in only 2\% of the observations, mistaking a change of scene for a change of task, whereas ours recognizes it in 66\% (\autoref{fig:teaser}). Over all ten tasks, the rates are 30\% and 74\%, and \autoref{fig:task_confusion} shows that LaST$_0$~\cite{liu2026last0} assigns most perturbed observations to the living-room tasks T5--T7, \ie, it identifies the scene rather than the task. Predicting isolated future frames thus binds the backbone to the appearance of the training scenes, whereas the motion-oriented derivation steers it toward what transfers across them.

\begin{figure}[t]
  \centering
  \includegraphics[width=\linewidth]{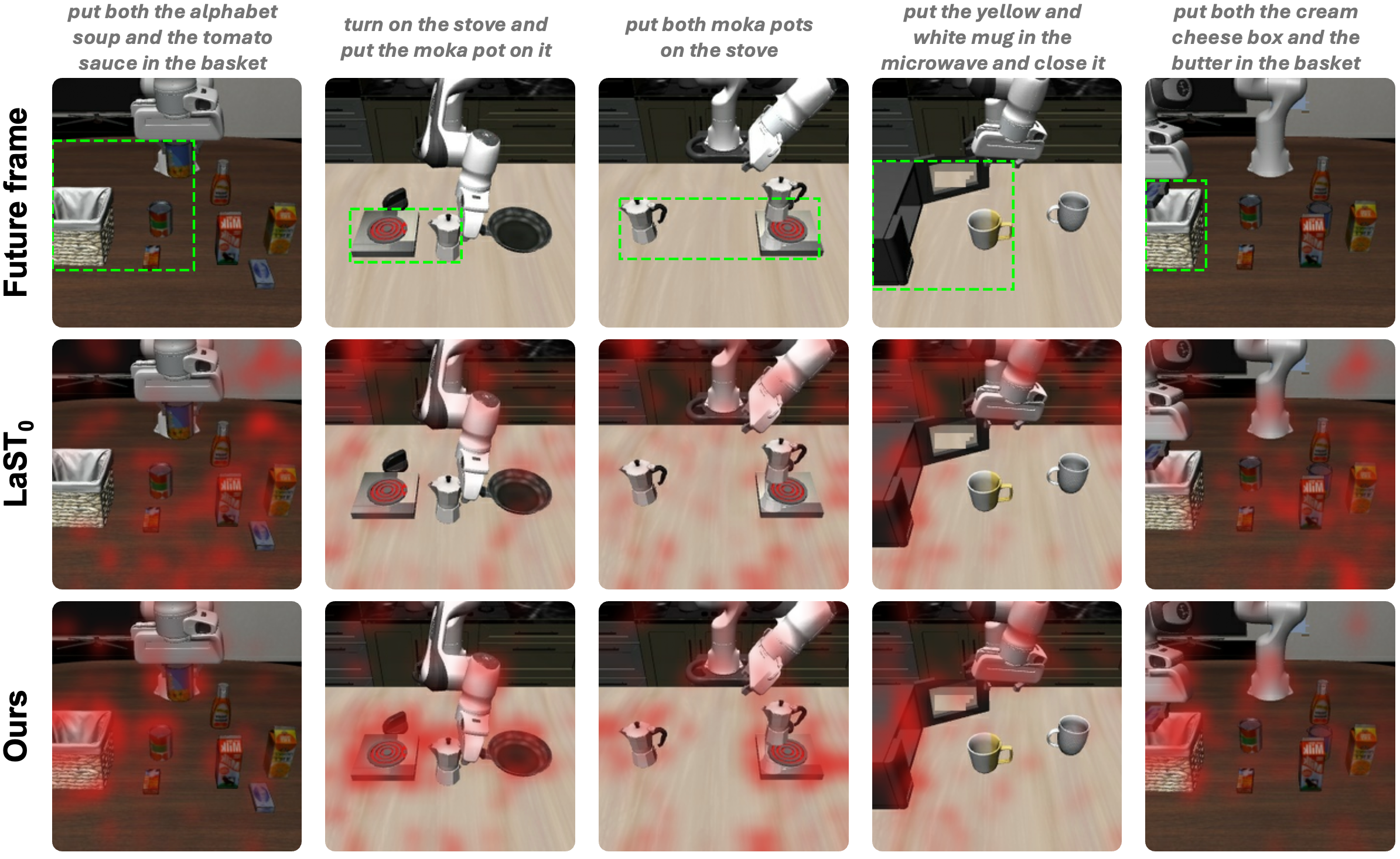}
  \caption{Similarity between the predicted latents and the patch features of the anticipated future frame on LIBERO-Long. Top: future frames with the annotated interaction region (dashed box). Middle: the visual latent of LaST$_0$~\cite{liu2026last0}. Bottom: the region latent of our method. Redder areas are more similar to the latent.}
  \vspace{-1.5em}
  \label{fig:region_vis}
\end{figure}

\noindent\textbf{What do the latents represent?}
We further inspect the predicted latents in the space where they are supervised. On 80 LIBERO-Long training states with annotated interaction regions, we compute the cosine similarity between each latent and the patch features of the future frame it anticipates. As shown in \autoref{fig:region_vis}, the snapshot latent of LaST$_0$~\cite{liu2026last0} resembles patches scattered over the scene, whereas our region latent concentrates on the interaction region. Over all 320 latent--frame pairs, 38.5\% of our top-10\% patches fall inside the region, against 15.7\% for LaST$_0$~\cite{liu2026last0} and 19.4\% by chance.

\subsection{Real-World Experiments}
\label{sec:exp_real}

\begin{figure*}[t]
  \centering
  \includegraphics[width=\textwidth]{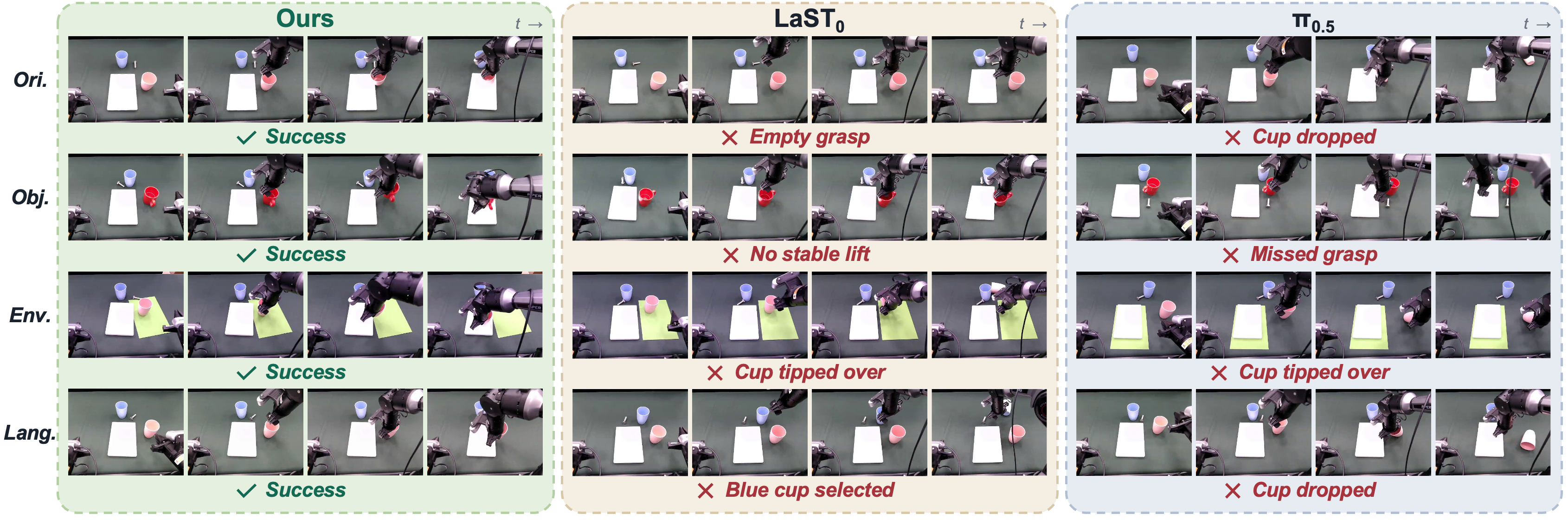}
  \vspace{-1.5em}
  \caption{Real-world rollouts on \textit{pick up the red cup} in the training scene (\textit{Ori.}) and under object (\textit{Obj.}), environment (\textit{Env.}), and language (\textit{Lang.}) perturbations. Our method succeeds in all four settings, whereas LaST$_0$~\cite{liu2026last0} and $\pi_{0.5}$ fail through empty, missed, or unstable grasps, drop or tip over the cup, or select the blue distractor cup. Frames are sampled uniformly over each episode from the main camera. Rollouts of our method on further tasks are provided in \autoref{app:real}.}
  \label{fig:real}
\end{figure*}

\begin{table}[t]
  \centering
  \caption{Success rates (\%) on an AgileX Piper dual-arm platform in distribution (ID), \ie, in the training scene, and under three out-of-distribution (OOD) perturbations: replaced objects or containers (Obj.), altered workspace backgrounds (Env.), and paraphrased instructions (Lang.). Numbers in brackets give the gain relative to LaST$_0$~\cite{liu2026last0}, and the best mean is in \protect\bestmark{bold}.}
  \label{tab:real}
  \setlength{\tabcolsep}{5pt}
  \resizebox{\linewidth}{!}{%
  \begin{tabular}{@{}llcccc}
    \toprule
    & & & \multicolumn{3}{c}{OOD} \\
    \cmidrule(lr){4-6}
    \multirow{-2}{*}{Tasks} & \multirow{-2}{*}{Methods} & \multirow{-2}{*}{ID} & Obj. & Env. & Lang. \\
    \midrule
    \rowcolor{sizeL}\cellcolor{white} & $\pi_{0.5}$~\cite{intelligence2025pi05} & 53.3 & 0.0 & 3.3 & 46.7 \\
    \rowcolor{sizeS}\cellcolor{white} & LaST$_0$~\cite{liu2026last0} & 73.3 & 23.3 & 6.7 & 73.3 \\
    \rowcolor{sizeS}\cellcolor{white} & \textbf{Ours} & 93.3 & 53.3 & 33.3 & 86.7 \\[-0.5pt]
    \rowcolor{sizeS}\cellcolor{white}\multirow{-4}{*}{\footnotesize\textit{Click the bell}} & & \gainrow{20.0} & \gainrow{30.0} & \gainrow{26.7} & \gainrow{13.3} \\
    \midrule
    \rowcolor{sizeL}\cellcolor{white} & $\pi_{0.5}$~\cite{intelligence2025pi05} & 66.7 & 63.3 & 3.3 & 46.7 \\
    \rowcolor{sizeS}\cellcolor{white} & LaST$_0$~\cite{liu2026last0} & 63.3 & 66.7 & 6.7 & 63.3 \\
    \rowcolor{sizeS}\cellcolor{white} & \textbf{Ours} & 96.7 & 93.3 & 36.7 & 96.7 \\[-0.5pt]
    \rowcolor{sizeS}\cellcolor{white}\multirow{-4}{*}{\footnotesize\textit{Pick up the red cup}} & & \gainrow{33.3} & \gainrow{26.7} & \gainrow{30.0} & \gainrow{33.3} \\
    \midrule
    \rowcolor{sizeL}\cellcolor{white} & $\pi_{0.5}$~\cite{intelligence2025pi05} & 66.7 & 56.7 & 0.0 & 43.3 \\
    \rowcolor{sizeS}\cellcolor{white} & LaST$_0$~\cite{liu2026last0} & 63.3 & 73.3 & 0.0 & 56.7 \\
    \rowcolor{sizeS}\cellcolor{white} & \textbf{Ours} & 96.7 & 96.7 & 43.3 & 96.7 \\[-0.5pt]
    \rowcolor{sizeS}\cellcolor{white}\multirow{-4}{*}{\footnotesize\textit{Pour the water}} & & \gainrow{33.3} & \gainrow{23.3} & \gainrow{43.3} & \gainrow{40.0} \\
    \midrule
    \rowcolor{sizeL}\cellcolor{white} & $\pi_{0.5}$~\cite{intelligence2025pi05} & 43.3 & 36.7 & 0.0 & 33.3 \\
    \rowcolor{sizeS}\cellcolor{white} & LaST$_0$~\cite{liu2026last0} & 43.3 & 46.7 & 13.3 & 16.7 \\
    \rowcolor{sizeS}\cellcolor{white} & \textbf{Ours} & 76.7 & 83.3 & 36.7 & 76.7 \\[-0.5pt]
    \rowcolor{sizeS}\cellcolor{white}\multirow{-4}{*}{\footnotesize\textit{Put blocks into cup}} & & \gainrow{33.3} & \gainrow{36.7} & \gainrow{23.3} & \gainrow{60.0} \\
    \midrule
    \rowcolor{sizeL}\cellcolor{white} & $\pi_{0.5}$~\cite{intelligence2025pi05} & 57.5 & 39.2 & 1.7 & 42.5 \\
    \rowcolor{sizeS}\cellcolor{white} & LaST$_0$~\cite{liu2026last0} & 60.8 & 52.5 & 6.7 & 52.5 \\
    \rowcolor{sizeS}\cellcolor{white} & \textbf{Ours} & \cellcolor{sizeSmean}\best{90.8} & \cellcolor{sizeSmean}\best{81.7} & \cellcolor{sizeSmean}\best{37.5} & \cellcolor{sizeSmean}\best{89.2} \\[-0.5pt]
    \rowcolor{sizeS}\cellcolor{white}\multirow{-4}{*}{\footnotesize Mean} & & \cellcolor{sizeSmean}\gainrow{30.0} & \cellcolor{sizeSmean}\gainrow{29.2} & \cellcolor{sizeSmean}\gainrow{30.8} & \cellcolor{sizeSmean}\gainrow{36.7} \\
    \bottomrule
  \end{tabular}}
\end{table}

\noindent\textbf{Data collection.}
We deploy our method on an AgileX Piper dual-arm
platform equipped with three external RGB cameras.
We record 14-dimensional joint-space states and
control targets covering both arms and grippers.
The policy uses the right arm only, predicting
7-dimensional end-effector actions comprising
translation, rotation, and gripper commands.
Further details on the action representation
and robot control are provided in \autoref{app:real_setup}.
We design four tabletop tasks: \textit{click the bell}, \textit{pick up the red cup} and place it in a white area with a blue distractor cup present, \textit{pour the water} from a kettle into a cup, and \textit{put blocks into cup}, which requires placing a red and a yellow cube into a cup. For each task, we collect approximately 60--90 teleoperated demonstrations.

\noindent\textbf{Training and evaluation details.}
We train task-specific policies using the
same architecture and two-pass training
strategy as in simulation, with
real-world-specific hyperparameters detailed
in \autoref{app:real_setup}, and compare them with LaST\(_0\)~\cite{liu2026last0} and \(\pi_{0.5}\)~\cite{intelligence2025pi05}. Beyond the training scene (ID), we evaluate three out-of-distribution (OOD) perturbations: \textit{Object} (\textit{Obj.}), which replaces the manipulated object or container, \eg, a silver bell with a red one or a handleless cup with a handled mug; \textit{Environment} (\textit{Env.}), which overlays colored background paper on the workspace; and \textit{Language} (\textit{Lang.}), which paraphrases the instruction, \eg, ``activate the bell by pressing it''. Specifically, object perturbations in the red-cup task replace the target cup while retaining the blue distractor cup and white placement area; in the pouring and block-placement tasks, they replace the receiving cup with a handled mug. Each task--condition pair is evaluated over 30 trials.

\noindent\textbf{Quantitative and qualitative analysis.}
As shown in \autoref{tab:real}, our method achieves a mean success rate of 90.8\% in the training scenes, surpassing LaST\(_0\)~\cite{liu2026last0} and \(\pi_{0.5}\) by 30.0 and 33.3 points, respectively. The advantage persists under perturbations: relative to LaST\(_0\)~\cite{liu2026last0}, our method improves the mean success rate from 52.5\% to 81.7\% under object changes, from 6.7\% to 37.5\% under environmental changes, and from 52.5\% to 89.2\% under paraphrased instructions. Selected qualitative examples illustrate these differences. When the original red cup is replaced with a handled mug, our method completes the transfer, whereas LaST\(_0\)~\cite{liu2026last0} fails to achieve a stable lift and \(\pi_{0.5}\) misses the grasp. The failure modes of the baselines are telling: given a paraphrased instruction, LaST\(_0\)~\cite{liu2026last0} grasps the blue distractor cup instead of the red one, and under environmental changes, both baselines topple the cup or fail to complete the transfer. These failures accord with the analysis in \autoref{sec:exp_ablation}: once the scene or instruction departs from training, a policy bound to the appearance of its training scenes falls back on memorized patterns, whereas our motion-oriented latents ground the action in the intended target (\autoref{fig:real}).

\section{Conclusion}
\label{sec:conclusion}

We have revisited the visual latents of latent VLAs, which inherit from Vision-Language Models the role of picturing an isolated future frame. Such snapshot targets bind the policy to the appearance of its training scenes. We have therefore proposed Motion-Oriented Latent Reasoning, which redirects the visual latents from the whole frame to the interaction that the action must bring about. Scaffolded by semantic latents that establish what to accomplish and how, these latents anticipate each future step from the global scene through the interaction region to the robot state. Our method improves on its latent counterpart both in distribution and under perturbations, in simulation and on a real robot. We hope these findings encourage latent reasoning for robotics to be designed around what the action must accomplish, not borrowed from image understanding.

\section*{Limitations}
\label{sec:limitations}

Our method relies on offline annotations, namely the stage descriptions of embodied chain-of-thought and the interaction regions obtained with open-vocabulary detection and segmentation, so that annotation errors carry over to the latent targets. Its region latent localizes the interaction of each future step with a single two-dimensional box, which cannot separate several interactions that occur at once, as in bimanual manipulation, and carries no depth, on which dexterous manipulation relies. Moreover, although the global and region targets are built from video frames without action labels, we have so far trained on robot demonstrations; learning them from large-scale human video, far more abundant than robot data, is left to future work.

\section*{LLM Usage Statement}
\label{sec:llm_statement}

We used Claude (Anthropic) and Codex (OpenAI), two large language model assistants, to edit the language of the manuscript and to format tables and \LaTeX{} source. The method, the experiments, and all scientific claims were conceived, conducted, and validated by the authors, who take full responsibility for the content of this paper.

{
    \small
    \bibliographystyle{ieeenat_fullname}
    \bibliography{main}
}

\clearpage
\setcounter{page}{1}
\maketitlesupplementary
\appendix

\section{Action Predictability}
\label{app:predictability}

This section details the oracle analysis in \autoref{sec:intro} and \autoref{fig:foresight}, which measures how much action-relevant information a set of features carries. On timesteps $t$ sampled from LIBERO-90 demonstrations, we compare four feature sets: the hidden states of the latent tokens of a Coconut-style~\cite{haotraining} latent ECoT~\cite{zawalski2024ecot,chen2025ecotlite} policy built on MiniVLA-1B (textual latents); the visual features of the current observation (initial state); the ground-truth visual features of the frames at $t{+}8$, $t{+}16$, $t{+}24$, and $t{+}32$ (visual foresight); and those of the same number of frames drawn from other timesteps of the same task (random frames). Each set is standardized, reduced to 64 dimensions by PCA, and fed to a ridge regression that predicts the ground-truth actions over the next 8 steps, with the sets concatenated when combined. Action predictability is the $R^2$ on the held-out 30\% of the episodes, averaged over action dimensions. The initial state alone attains 0.523; adding textual latents raises it to 0.616, whereas adding visual foresight raises it to 0.678. Replacing the foresight with random frames yields only 0.610, so the gain stems from the correspondence to the actual future rather than from the additional appearance.

\section{Real-World Experimental Setup}
\label{app:real_setup}

\begin{figure*}[t]
  \centering
  \includegraphics[width=\textwidth]{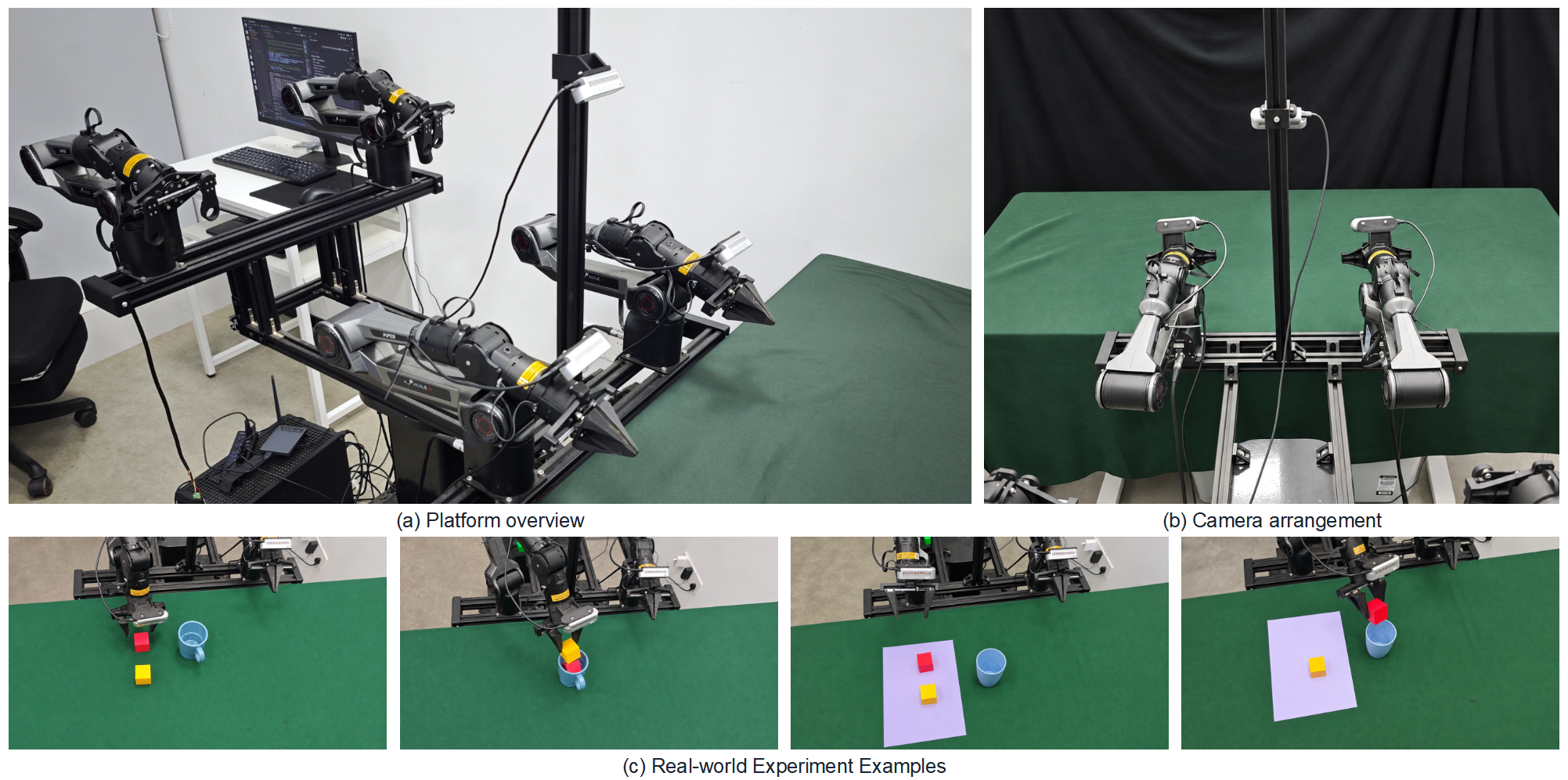}
  \caption{Real-world experimental setup with an AgileX Piper dual-arm platform
  and three external RGB cameras.}
  \label{fig:real_robot_setup}
\end{figure*}

This section describes the data collection, annotation, and deployment
procedures for the four real-world tasks in the main paper.
\autoref{fig:real_robot_setup} presents the real-world experimental setup.

\paragraph{Platform and data collection.}
We use an AgileX Piper dual-arm platform with three external RGB cameras and
collect demonstrations through leader--follower teleoperation.
Each arm has six joints and one gripper channel.
At each time step, we record the three RGB views, joint and gripper
feedback, control targets, and the task instruction.
The recordings use a sampling rate of $30\,\mathrm{Hz}$ and an image
resolution of $640\times480$.
Visual observations and numerical records are associated through
episode indices, frame indices, and timestamps, and stored in the
LeRobot format.
We collect approximately 60–90 teleoperated demonstrations for each of the four tasks: bell pressing, red-cup pickup, water pouring, and block placement.

\paragraph{State and action representation.}
The raw recordings contain 14-dimensional states and joint-space control targets for both arms. Our policy uses the right arm, with a seven-dimensional state comprising six joint angles in radians and the gripper opening in meters. We convert the recorded control targets into end-effector action labels using Piper forward kinematics. Let $(p_t,R_t)$ and $(p_t^{\mathrm{tar}},R_t^{\mathrm{tar}})$ denote the end-effector poses corresponding to the current feedback and the recorded target, respectively. The action is
\begin{equation}
 a_t=\left[
 \Delta p_t^{\mathsf T},\,
 \Delta r_t^{\mathsf T},\,g_t
 \right]^{\mathsf T}\in\mathbb{R}^{7},
 \label{eq:real_action}
\end{equation}
where $\Delta p_t=p_t^{\mathrm{tar}}-p_t$ is expressed in the right-arm base frame, and $\Delta r_t=\operatorname{Euler}_{XYZ}(R_t^{\mathrm{tar}}R_t^{\mathsf T})$ uses extrinsic XYZ angles in radians. The absolute gripper target $g_t\in[0,1]$ represents the opening fraction over the configured $0$--$0.08\,\mathrm{m}$ range. Thus, action labels describe the displacement from current feedback to the control target, rather than the displacement between consecutive feedback frames. The bell task keeps the gripper fixed and masks its action channel from supervision.

\paragraph{Temporal supervision.}
Each sample contains a 16-step action chunk starting at the current frame. Future visual and proprioceptive targets are taken at offsets $\{8,16,24,32\}$, corresponding to approximately $\{0.27,0.53,0.80,1.07\}$ seconds. All three RGB views are used as current visual inputs to the policy, and future visual targets are taken from the main camera. Targets extending beyond an episode are padded with its final frame or action. When no valid interaction-region box is available, including for future frames beyond the episode boundary, the regional target falls back to the global visual target. No supervision crosses episode boundaries.

\paragraph{Reasoning and spatial annotations.}
We organize offline reasoning supervision into five fields: \textit{Subtask}, \textit{Why}, \textit{Movement}, \textit{Movement reasoning}, and \textit{Plan}. These describe the current operation, its purpose, the intended motion, the motion rationale, and the remaining steps. Descriptions are conditioned on the task, the manipulation phase, and the recorded action, with visual observations and robot motion providing the context. Movement descriptions refer to the current control target, and gripper descriptions distinguish closing, holding, and releasing. For spatial supervision, we associate task-relevant objects with their interaction or placement regions in the main view. When both object and destination boxes are available, their enclosing rectangle defines the region of interest. Box coordinates are normalized by image width and height and matched to the corresponding future frame. When no valid interaction-region box is
available, the regional target falls back
to the corresponding global visual target. These textual and spatial annotations provide offline supervision and are not required as inputs during deployment.

\paragraph{Training.}
We train a separate policy for each real-world task. We initialize the
vision--language backbone with Janus-Pro-1B weights and the action expert
with pretrained weights from LaST$_0$~\cite{liu2026last0}. The visual encoders and visual projection module remain frozen, while the language and action components are fine-tuned. Each policy uses 17 latent tokens: five reasoning tokens followed by four groups of global visual, ROI visual, and proprioceptive tokens, corresponding to the future offsets defined above. The training objective combines a masked flow-matching action loss, cosine alignment of future visual and proprioceptive latents, reasoning-embedding alignment, and autoregressive text reconstruction. The latent alignment objectives are supervised
in the first pass using teacher-forced inputs, while the action loss in the second pass is conditioned on the latents predicted in the first pass, rather than on ground-truth latent targets.

Demonstrations are split by episode into approximately $90\%$ training and $10\%$ validation data using seed 42. Joint angles and the six end-effector displacement components are normalized to $[-1,1]$ using the 1st and 99th percentiles computed exclusively from the training episodes. We optimize with AdamW on eight GPUs using bfloat16 precision, a peak learning rate of $5\times10^{-5}$ for both the language and action components, linear warmup over the first $5\%$ of training steps, and cosine decay to zero. Weight decay is disabled, and the gradient norm is clipped to 1.0. 

\paragraph{Closed-loop deployment.}
At inference time, the policy receives the task instruction, the three current RGB views, and the right-arm state, using the same view ordering, units, and normalization statistics as during training. The flow-based action expert generates a 16-step chunk with 10 integration steps. We execute the first 8 actions before acquiring fresh observations and predicting the next chunk. Execution uses a nominal $30\,\mathrm{Hz}$ control loop; this is the command rate rather than the rate of full policy inference. Before each command, the predicted action is denormalized and applied to the latest measured end-effector pose. Joint-limit-constrained numerical inverse kinematics, initialized from the current joint feedback, converts the target pose into joint commands. Per-command joint changes and cumulative displacement within a chunk are bounded. Infeasible inverse-kinematics solutions trigger a hold, with repeated failures stopping execution. The left arm is not commanded. The right gripper follows the predicted opening for grasping tasks and remains fixed for the bell task.

\paragraph{Evaluation protocol.}
Each model is evaluated on the four real-world tasks in the training scene and under object, environment, and language perturbations. For each model, 30 trials are conducted for every task–condition pair, and success rates are computed separately for each pair.

\section{Interaction-Region Annotation}
\label{app:bbox}

The region latent (\autoref{sec:visual_latents}) requires, for every training frame, the box $\mathcal{B}_j$ enclosing the manipulated object and its destination. We construct these boxes offline as follows; they serve only to build the training targets and are never used at inference.

\noindent\textbf{Object tracks.} For the manipulated object, per-frame box tracks are obtained by prompting SAM~3~\cite{carion2025sam3} with Grounding DINO~\cite{liu2024groundingdino} detections and are verified step by step.

\noindent\textbf{Destinations.} Destinations remain static within an episode, \eg, the plate in LIBERO-Spatial and the basket in LIBERO-Object. We therefore detect each destination once, on the first frame of the episode, with Grounding DINO (Swin-T) prompted by the name of the destination, using box and text thresholds of $0.25$ and $0.20$ and non-maximum suppression at an IoU of $0.5$, and reuse the box for all frames. On BridgeData~V2, the destination phrase is parsed from the instruction; when it names a location rather than an object, \eg, \textit{the lower right corner}, the corresponding region of the workspace serves as the destination.

\noindent\textbf{Gripper.} For operations without grasping, such as turning on the stove or opening a drawer, the gripper takes the place of the object. We detect it with Grounding DINO, prompted by \textit{white gripper on robot arm} with a box threshold of $0.12$, and track it with SAM~3 prompted by the detected box. A box prompt is preferred over a point prompt, which tends to expand the mask to the entire arm.

\noindent\textbf{Merging and quality control.} The interaction region of each frame is the tightest box enclosing its component boxes, expressed on the $224{\times}224$ input image. We visualize the boxes of every episode and correct failures manually or by re-detection with revised prompts, \eg, missed or duplicated destinations and unparsable destination phrases. Frames without a valid box, including future frames beyond the end of an episode, fall back to the global target $\bar{\mathbf{g}}_j$.

\section{Additional Real-World Rollouts}
\label{app:real}

\begin{figure*}[t]
  \centering
  \includegraphics[width=\textwidth]{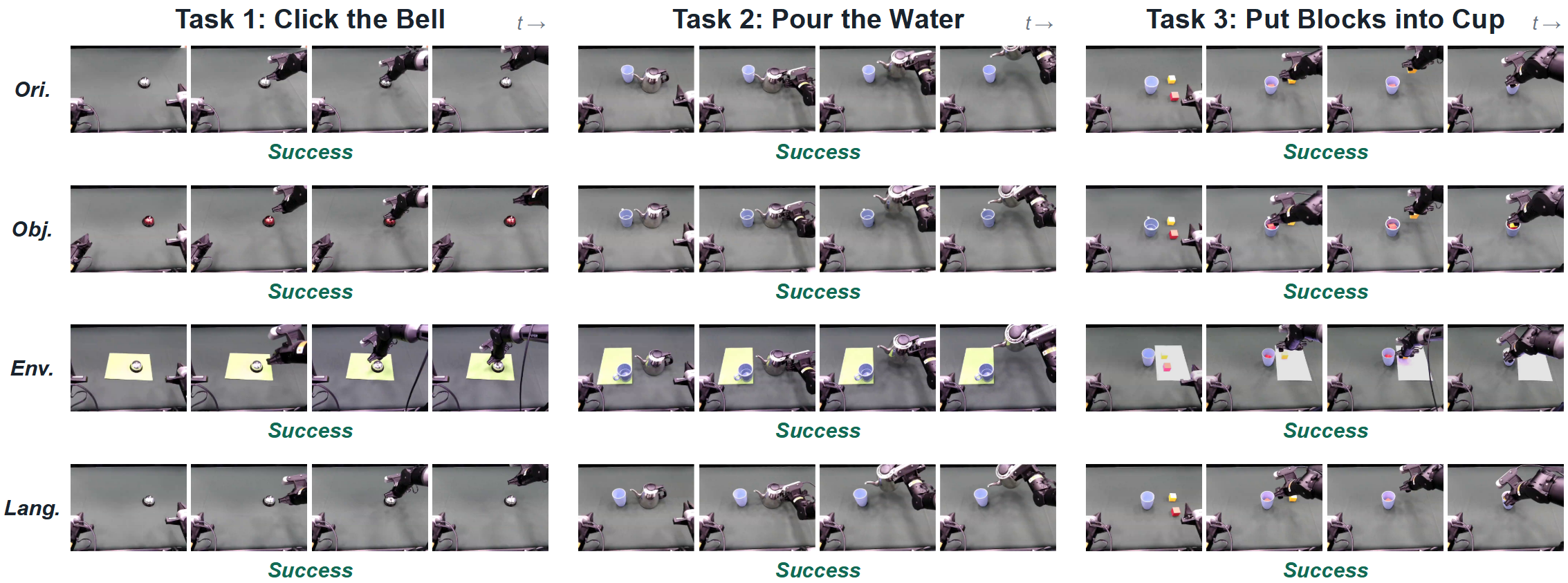}
  \caption{Real-world rollouts of our method in the training scene (\textit{Ori.}) and under object (\textit{Obj.}), environment (\textit{Env.}), and language (\textit{Lang.}) perturbations. Frames are sampled uniformly over each episode from the main camera.}
  \label{fig:real_more}
\end{figure*}

\autoref{fig:real_more} presents rollouts of our method on \textit{click the bell}, \textit{pour the water}, and \textit{put blocks into cup}, complementing the comparison in \autoref{fig:real}.

\end{document}